\documentclass[runningheads]{llncs}
\usepackage{graphicx}
\usepackage{url}
\usepackage{cite}
\newcommand\blfootnote[1]{%
  \begingroup
  \renewcommand\thefootnote{}\footnote{#1}%
  \addtocounter{footnote}{-1}%
  \endgroup
}

\begin{document}
\title{Automatic identification of segmentation errors for radiotherapy using geometric learning}
\titlerunning{Automatic identification of segmentation errors using geometric learning}
%
\author{Edward G. A. Henderson\inst{1} \and
Andrew F. Green\inst{1,2} \and
Marcel van Herk\inst{1,2} \and
Eliana M. Vasquez Osorio\inst{1,2}}
%
\authorrunning{E. G. A. Henderson et al.}
%
\institute{The University of Manchester, Oxford Rd, Manchester M13 9PL, UK \and Radiotherapy Related Research, The Christie NHS Foundation Trust, Manchester M20 4BX, UK\\
\email{edward.henderson@postgrad.manchester.ac.uk}\\
}
\maketitle              
\begin{abstract}
Automatic segmentation of organs-at-risk (OARs) in CT scans using convolutional neural networks (CNNs) is being introduced into the radiotherapy workflow. However, these segmentations still require manual editing and approval by clinicians prior to clinical use, which can be time consuming. The aim of this work was to develop a tool to automatically identify errors in 3D OAR segmentations without a ground truth. Our tool uses a novel architecture combining a CNN and graph neural network (GNN) to leverage the segmentation’s appearance and shape. The proposed model is trained using self-supervised learning using a synthetically-generated dataset of segmentations of the parotid gland with realistic contouring errors. The effectiveness of our model is assessed with ablation tests, evaluating the efficacy of different portions of the architecture as well as the use of transfer learning from an unsupervised pretext task. Our best performing model predicted errors on the parotid gland with a precision of 85.0\% \& 89.7\% for internal and external errors respectively, and recall of 66.5\% \& 68.6\%. This offline QA tool could be used in the clinical pathway, potentially decreasing the time clinicians spend correcting contours by detecting regions which require their attention. All our code is publicly available at \url{https://github.com/rrr-uom-projects/contour_auto_QATool}.
\end{abstract}
\blfootnote{\textbf{Accepted in 25th International Conference on Medical Image
Computing and Computer Assisted Intervention (MICCAI 2022)}}
\vspace{-5mm}
\section{Introduction}
The 3D segmentation of organs-at-risk (OARs) in computed tomography (CT) scans is a crucial step in the radiotherapy pathway. However, manual segmentation is time-consuming and prone to inter- and intra-observer variability\cite{Brouwer2012}. Automatic segmentation using convolutional neural networks (CNNs) is now considered state-of-the-art for medical image segmentation\cite{Cardenas2019}. However, every contour needs to be evaluated and approved by clinical staff before use in treatment planning\cite{Vandewinckele2020a}. The aim of this study was to develop a quality assurance tool to automatically detect segmentation errors without a ground-truth. Such a tool could complement current radiotherapy workflows by drawing the clinicians' attention to regions of a contour which may need updating.

There are already a few methods for automatic quality-assurance (QA) of anatomical segmentations. Some approaches use statistical models to classify contours requiring further updates or to predict performance metrics. In 2012, Kohlberger et al. applied a support vector machine regression algorithm to predict a volumetric overlap metric for segmentations based on hand-selected shape and appearance features\cite{Kohlberger2012}. Chen et al. trained a statistical model based on geometrical attributes of segmentations and their neighbouring structures which could classify contours as correct or incorrect\cite{Chen2015}. McCarroll et al. used a bagged tree classification model to flag contours needing attention\cite{mccarroll2017machine}, while Hui et al. developed a QA tool using a hand-crafted set of volumetric features\cite{Hui2018}.

Some more recent studies used secondary auto-segmentation methods or CNN model ensembling to highlight regions of uncertainty in auto-contouring\cite{Rhee2019, Men2020, Mehrtash2020}. Valindria et al. proposed a reverse classification accuracy method in which a secondary segmentation model was trained from an image and segmentation prediction pair without a ground truth\cite{Valindria2017}. By evaluating this new model on a database of images with known ground truth segmentations they could predict the Dice similarity coefficient (DSC) for the original prediction. Sander et al. developed a two-step auto-segmentation method for cardiac MR images capable of estimating uncertainties and predicting local segmentation failures\cite{Sander2020}. 

In contrast, our proposed automatic QA method can be applied to any segmentation, automatic or manual, and estimates error on the entire 3D segmentation shape without a gold-standard. In this proof of concept study we predicted errors on a single OAR delineated in head and neck CT scans. We chose the parotid gland as it is suitably complex, with both convex and concave regions, to showcase a wide range of potential segmentation failures. 

In this paper we make several contributions. First, we develop a novel automatic error detection system, combining convolutional neural networks (CNNs) and Graph Neural Networks (GNNs). We perform ablation studies to individually test different portions of our method. We also perform pre-training and transfer learning to speed up training convergence. Our model is self-supervised in the sense that unlimited training data can be generated from a small amount of OAR segmentations and images prior to training.

\section{Materials and method}
Our proposed automatic QA tool uses multiple forms of input information: shape information of the segmentation structure and information from the appearance of the surrounding CT scan. To train our method we used a public dataset with segmentations and generated additional segmentations with realistic errors.

\subsection{Dataset}
For this study we used a publicly-available dataset of 34 head and neck (HN) CT scans\cite{nikolov2018}. Each of the 34 HN CTs have segmentations of OARs performed by two doctors. For this study we used one doctor's segmentations of the left and right parotid glands as the ground truth. For a given patient, the left and right parotid glands are bilateral structures and mostly symmetric. However, between patients the gland shapes vary considerably. All patient data was flipped laterally to create a dataset of 68 left parotid gland ground truth segmentations.

\subsection{Generating a training dataset}
\label{ssec:dataset}
\begin{figure}[ht]
\includegraphics[width=\textwidth]{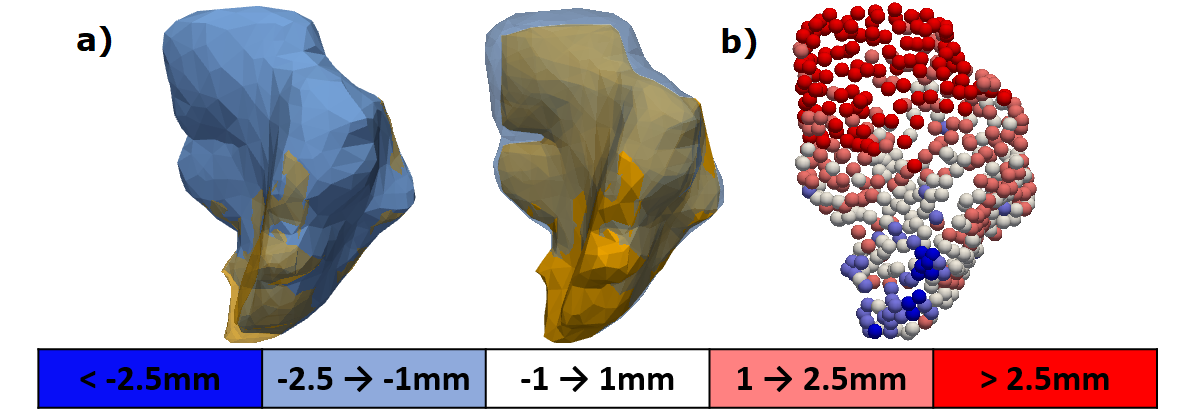}
\caption{An example of a perturbed segmentation and classes related to the distance from the ground truth. The ground-truth segmentation is shown in orange and the perturbed segmentation is shown in blue. The coloured nodes on the right correspond to the signed closest distance to the ground truth.}\label{perturbs_fig}
\end{figure}
\vspace{-5mm}
To create a large dataset of training samples containing segmentation errors we perturbed the ground truth parotid gland segmentations (Fig. \ref{perturbs_fig}a). Practically, we created signed distance transforms of the ground-truth segmentations and added structured noise voxel-wise. The structured noise was created by drawing noise from a normal distribution which was then convolved with a 7.5mm Gaussian kernel with an amplitude such that the resultant structured noise had a standard deviation of $\sim$1mm. The perturbed segmentation was extracted using the marching cubes algorithm (at level=0) to acquire a triangular mesh manifold.  A connected components algorithm was applied to eliminate any disconnected spurious segmentations. To simplify the mesh we applied 100 iterations of Taubin smoothing\cite{Taubin1995} and quadric error metric decimation reducing the mesh size to $\sim1000$ triangles followed by a final 10 iterations of Taubin smoothing. We perturbed each ground-truth segmentation 100 times to generate a dataset of 6800 parotid glands. 

Node-wise classes were then determined for each of the perturbed segmentations. The signed nearest distance to the ground truth segmentation was calculated for each node, categorising them into one of five classes (Fig. \ref{perturbs_fig}b).


\subsection{A hybrid CNN-GNN model for contour error prediction}

\begin{figure}[ht]
\includegraphics[width=\textwidth]{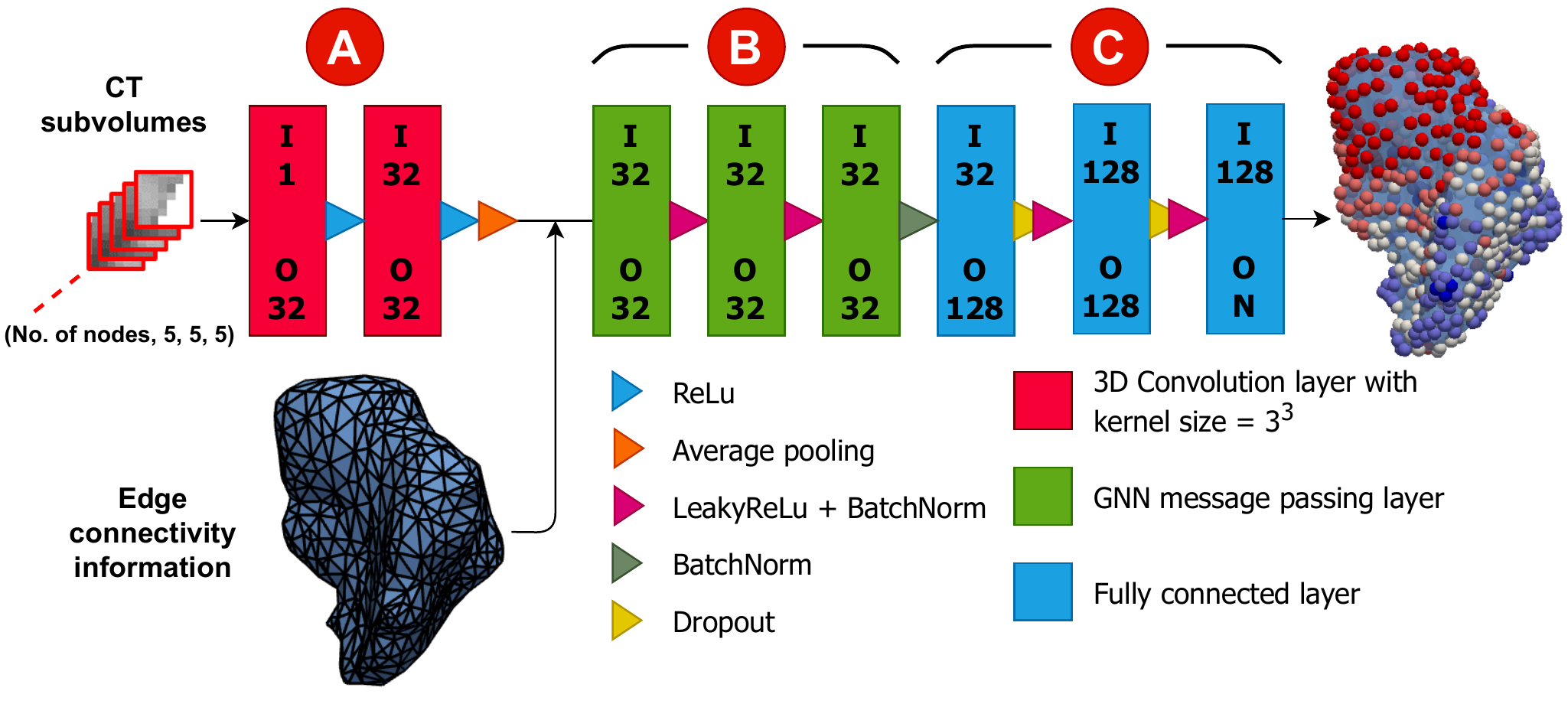}
\caption{An illustration of our CNN-GNN architecture. A small 3D patch for each node was extracted from the CT scan and provided as input to the CNN encoder (\textbf{A}). The edge connectivity information for the segmentation structure is provided to the GNN in order to perform message passing (\textbf{B}). The MLP decoder is formed of 3 fully-connected layers which make the node-wise classification predictions (\textbf{C}). The output on the right is an actual prediction.} \label{CNNGNN_fig}
\end{figure}

Our proposed method performs node-wise classification and makes predictions which correspond to the signed distance from the ground-truth segmentation without explicit knowledge of the ground-truth (Fig. \ref{perturbs_fig}b). Since we implemented a novel method we investigated several architectures, considering different types of GNN message passing layers to use in the processor and the style of feature extractor to use for the CNN encoder. For the sake of conciseness in this paper we only describe our best architecture with brief justifications.

We used an encoder, processor and decoder structure inspired by the work of Pfaff et al. on learning mesh-based simulations using GNNs\cite{Pfaff2021}. Figure \ref{CNNGNN_fig} shows a schematic of our model structure, which consists of a CNN encoder, a GNN processor and a multi-layer perceptron (MLP) decoder (Fig. \ref{CNNGNN_fig}, labels \textbf{A}, \textbf{B} and \textbf{C}, respectively). A $5\times5\times5$ voxel CT sub-volume centred on each node was provided to the CNN encoder, a feature extractor of two 3D convolutional layers, to produce node-wise representations. The GNN takes these representations and performs message passing between connected nodes, updating each node's representation according to its local neighbourhood. The MLP takes the final representations to predict node-wise classifications. Our hypotheses behind this model design were that the CNN encoder would allow our method to gain context from the CT scan appearance while the GNN processor would leverage the geometrical structure of the input and make node-wise predictions with awareness of the local neighbourhood, similar to active appearance models\cite{cootes1998active}.

For the GNN processor we used the \textit{SplineConv} layer, introduced by Fey et al., to perform message passing\cite{Fey2018}. The \textit{SplineConv} is a generalisation of traditional CNN convolution layers for geometric input. Node features in local neighbourhoods are weighted by B-Spline continuous kernel functions, which are in turn parameterised by sets of trainable weights. How node features are aggregated among neighbouring nodes is determined by spatial relations. The relative spatial positions of each node are encoded using 3D Cartesian pseudo-coordinates on the range [0,1]. In our GNN layers we used the addition aggregation operation as this has been shown to produce the most expressive GNN models\cite{Xu2019a}. All \textit{SplineConv} layers in our model used spline degree 2, kernel size 5 and were followed by LeakyReLU activation functions and batch normalisation.

\subsection{Unsupervised pre-training}
\label{ssec:pretraining}
\begin{figure}[ht]
\includegraphics[width=\textwidth]{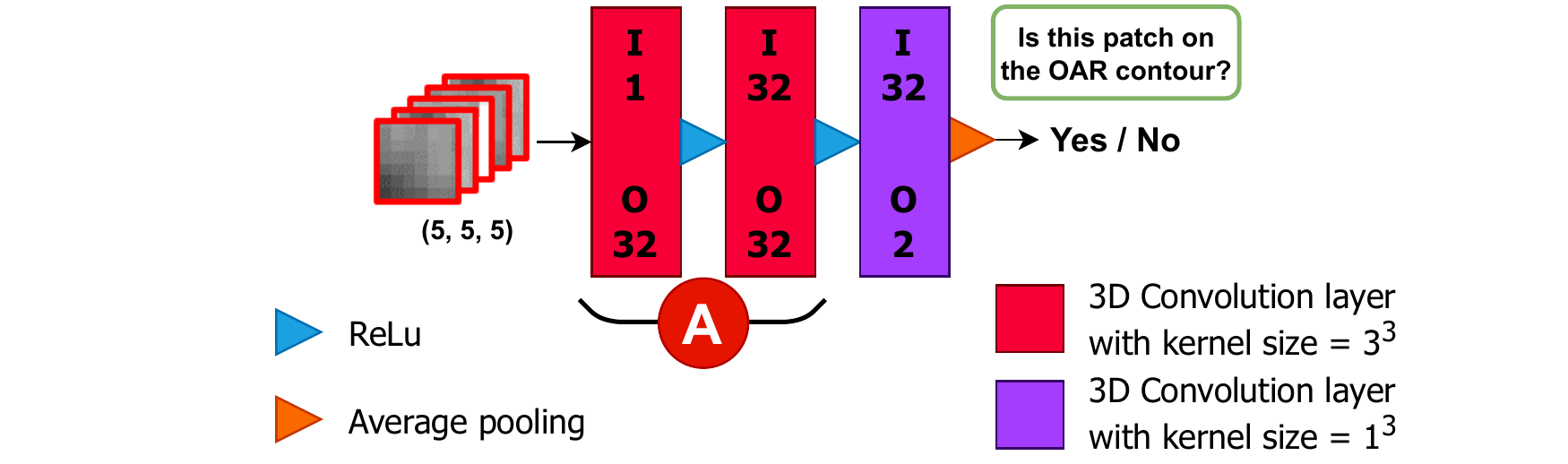}
\caption{A schematic of the pre-training task. The CNN encoder, marked A, was connected to a 1x1x1 convolution layer and an average pooling operation to perform binary classification. The pretext task consisted of classifying whether 5x5x5 CT patches were sampled from the parotid gland boundary.} \label{pretraining_fig}
\end{figure}


Pre-training is a popular transfer learning technique where a model is trained to perform a pretext task in order to learn a set of features which are helpful for the model when applied to the main task\cite{Noroozi2016}. We applied it to initialise the CNN encoder of our model.

We designed a custom self-supervised pretext task: to classify 3D CT patches as either ``on'' or ``off'' the parotid gland boundary (Fig. \ref{pretraining_fig}). These 3D CT image patches were extracted from the same dataset described in section \ref{ssec:dataset} live at training time in an equal ratio of ``on'' and ``off'' contour.

\subsection{Implementation details}
All our models are implemented using PyTorch 1.9.0 and PyG 2.0.1\cite{Fey/Lenssen/2019} and operations with triangular meshes are performed using Open3D 0.13.0\cite{Zhou2018}. Our proposed model has 437,829 parameters. All model training was performed using a 24GB NVidia GeForce RTX 3090 and AMD Ryzen 9 3950X 16-Core Processor.

In the pre-training phase the encoder was trained on 512 patch samples per epoch with a batch size of 64 for a maximum of 10000 epochs. Our error-prediction models were trained on 25 perturbations per image per epoch, with a batch size of 16 for a maximum of 50 epochs. For both phases we used the AdamW optimizer\cite{loshchilov2019} with an initial learning rate of 0.001, weight decay set to 0.001 and learning rate cosine annealing. Training individual models took $\sim10$ minutes and inference for a single segmentation takes $\sim 0.03$ seconds.

\section{Experiments}
We performed a series of experiments in which we ablated different portions of our proposed method. A 5-fold cross validation was performed for each experiment. In each fold, we used 4800 perturbed structures for training and 600 for validation. Sets of 1400 testing structures were held out and used to evaluate the final classification performance of the fold. Since we used the same dataset for both the pre-training and error-prediction tasks the same cross-validation seeding was used for each. This ensured the same images and contours were used for training, validation and testing in each fold between the two tasks.

\subsection{Ablation tests}
First, to determine the value of the CNN encoder we trained an identical set of models with the 3D CT sub-volumes uniformly overwritten with a constant value. This effectively blinded our model to the CT scan appearance. With this ablation the model learned to predict errors from the segmentation shape alone. Second, to evaluate whether the GNN processor was learning useful information from the segmentation's geometric structure, we trained a model without GNN message passing. In this ablation the node-wise representations from the encoder were passed directly to the MLP decoder. Finally we trained an additional set of models from scratch to analyse whether pre-training the CNN encoder, as described in section \ref{ssec:pretraining}, was advantageous.

Confusion matrices were used to evaluate these tests. We calculated each models precision and recall of internal and external segmentation errors (predictions of the edge classes, $<-2.5$mm and $>2.5$mm). Errors predicted with the correct sign were considered correct for precision but not recall.

\section{Results}
\begin{figure}[t]
\includegraphics[width=\textwidth]{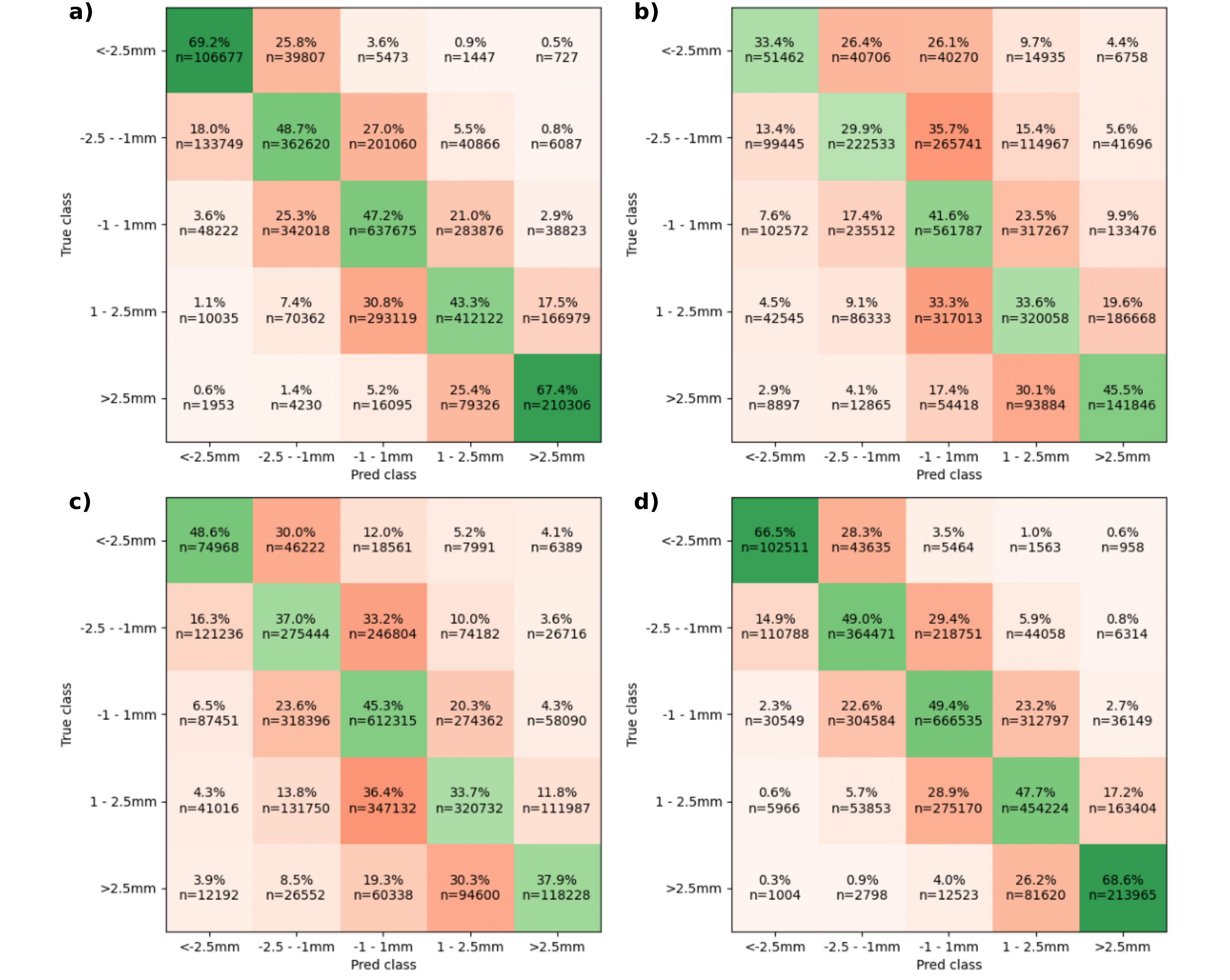}
\caption{Confusion matrices of the results of our experiments. a) Our proposed CNN-GNN-MLP; b) CNN encoder ablation; c) GNN processor ablation; d) pre-training ablation. The percentages in each plot corresponds to the recall.}\label{results_fig}
\end{figure}

In Figure \ref{results_fig} we show confusion matrices of the results of our proposed method (Fig. \ref{results_fig}a) alongside the results of the 3 experiments performed (Fig. \ref{results_fig}b-d). The percentages included in the confusion matrices show the recall for each class. The internal and external precision for our full model with pre-training was $79.9\%$ and $89.2\%$.

\subsection{Ablation tests}
By comparing Fig.\ref{results_fig}b and Fig.\ref{results_fig}a we observed that the CNN encoder was necessary for our method to make good predictions, i.e. the CT input was essential. This is unsurprising as the segmentation under consideration could have identical shape to the ground truth but a spatial offset would introduce errors. Information from the imaging is the only way of determining if this is the case.

Comparing Fig. \ref{results_fig}c and Fig.\ref{results_fig}a shows that without the GNN the models prediction performance decreased with the internal and external recall dropping from $69.2\%$ to $48.6\%$ and $67.4\%$ to $37.9\%$. The internal and external precision for the GNN-ablated model dropped to $58.2\%$ and $71.6\%$. The GNN shared information between connected neighbours. As a result, the GNN processor smoothed predictions made in local regions with a higher proportion of nodes allocated similar classes (i.e. $\pm1$ class) as their neighbours ($0.974\pm0.012$) compared to the GNN-ablated model ($0.894\pm0.024$). The proportion of matching neighbours for the ground truth was $0.967\pm0.014$. A figure of the full distributions is included in the supplementary materials (Fig. S.1).

The mean accuracy over the five-fold cross-validation for the pretext task was $88.8\pm0.7\%$. However, by comparing Fig.\ref{results_fig}d and Fig.\ref{results_fig}a we observed that pre-training had little positive impact on the final error-prediction rates. The internal and external precision for the full model without pre-training was actually best at $85.0\%$ and $89.7\%$. Models with a pre-trained encoder had less erratic validation loss during training, but converged to the similar loss values. The complete validation loss curves are included in the supplementary materials (Fig. S.2).

\section{Discussion}

In this study we developed a method to identify errors in OAR segmentations from radiotherapy planning CT scans without a ground truth. Our proposed model leverages both the CT image appearance and the segmentation's shape with a novel hybrid CNN-GNN-MLP architecture. To the best of our knowledge this is the first time geometric learning with GNNs has been applied to segmentation error prediction. Using GNNs, our method gained greater context of the geometric shape of the segmentation by leveraging the manifold structure directly. In our study, we found that pre-training the CNN part of the model did not improve prediction results, but only enabled a smoother training process.

Several approaches have been published to perform automatic segmentation QA based on statistical models or using multiple segmentation models. However, most of these previous methods predict a global accuracy measure such as DSC\cite{Kohlberger2012, Chen2015, mccarroll2017machine, Valindria2017, Rhee2019, Men2020}. Our proposed method makes node-wise predictions of errors in distinct locations on the segmentation to help guide clinicians when completing the manual contour QA process.

There have been some previous works to predict uncertainties using model ensembles\cite{Mehrtash2020} or directly using new CNN models\cite{Sander2020}. However, these approaches require the adoption of the new segmentation models themselves. Our method can be applied to segmentations generated by any means, allowing QA of existing methods. In particular, our proposed method fills a void as most current commercial auto-segmentation packages do not provide uncertainty estimations\cite{Green2022}.

The perturbed segmentations that were generated for the training dataset were not informed by real anatomical shape variations at this time. In future, training data could be generated based on observer variation data\cite{vasquez_osorio2019}.

Additionally, in this study we only performed experiments on one OAR, the parotid gland, in the head and neck anatomy. This is a difficult organ to segment and creates a broad range of error scenarios for our model to learn to recognise. Given our proposed methods performance on this OAR, we are confident our method can be extended to other anatomical structures in the future.

As it stands, our tool could easily be dropped into the clinical pathway, potentially decreasing the time clinicians spend correcting contours by drawing their attention to regions that need updating. It is potentially also fast enough to be applied in an interactive setting, e.g. during the final editing process.

\section{Conclusion}
We have developed a novel method to predict segmentation errors without a ground-truth. Our model combines shape and appearance features using a hybrid CNN-GNN architecture. This method could provide automatic segmentation QA to improve consistency and outcomes for patients treated with radiotherapy.

\small{\subsubsection{Acknowledgements.}
Marcel van Herk was supported by NIHR Manchester Biomedical Research Centre. This work was also supported by Cancer Research UK via funding to the Cancer Research Manchester Centre [C147/A25254] and by Cancer Research UK RadNet Manchester [C1994/A28701].}

%
%
%
\bibliographystyle{splncs04}
\bibliography{bibliography}

\end{document}